\documentclass[runningheads]{llncs}

\usepackage{eccv}

\usepackage{eccvabbrv}

\usepackage{graphicx}
\usepackage{booktabs}

\usepackage{multirow}
\usepackage{color, colortbl}

\usepackage[symbol]{footmisc}
\definecolor{Gray}{gray}{0.9}

\usepackage[accsupp]{axessibility}  % Improves PDF readability for those with disabilities.

\usepackage[pagebackref,breaklinks,colorlinks]{hyperref}
\usepackage{orcidlink}

\begin{document}

\title{LabelDistill: Label-guided Cross-modal Knowledge Distillation for Camera-based 3D Object Detection} 

\titlerunning{LabelDistill: Label-guided Cross-modal Knowledge Distillation for 3DOD}

\author{Sanmin Kim\inst{1}\orcidlink{0000-0002-4042-6570} \and
Youngseok Kim\inst{2}\thanks{Work done at Korea Advanced Institute of Science and Technology.}\orcidlink{0000-0001-9984-2416} \and
Sihwan Hwang\inst{1}\orcidlink{0009-0003-1557-4664} \and
Hyeonjun Jeong\inst{1}\orcidlink{0009-0003-2679-3280} \and
Dongsuk Kum\inst{1}\orcidlink{0000-0002-2590-4845}}

\authorrunning{Kim et al.}

\institute{Korea Advanced Institue of Science and Technology, Daejeon, South Korea\\
\email{\{sanmin.kim, shhwang0129, hyeonjun.jeong, dskum\}@kaist.ac.kr} \and
42dot Inc., Seoul, South Korea\\
\email{youngseok.kim@42dot.ai}
}

\maketitle

\begin{abstract}
%New abstract
Recent advancements in camera-based 3D object detection have introduced cross-modal knowledge distillation to bridge the performance gap with LiDAR 3D detectors, leveraging the precise geometric information in LiDAR point clouds. 
However, existing cross-modal knowledge distillation methods tend to overlook the inherent imperfections of LiDAR, such as the ambiguity of measurements on distant or occluded objects, which should not be transferred to the image detector.
To mitigate these imperfections in LiDAR teacher, we propose a novel method that leverages aleatoric uncertainty-free features from ground truth labels.
In contrast to conventional label guidance approaches, we approximate the inverse function of the teacher's head to effectively embed label inputs into feature space.
This approach provides additional accurate guidance alongside LiDAR teacher, thereby boosting the performance of the image detector.
Additionally, we introduce feature partitioning, which effectively transfers knowledge from the teacher modality while preserving the distinctive features of the student, thereby maximizing the potential of both modalities.
Experimental results demonstrate that our approach improves mAP and NDS by 5.1 points and 4.9 points compared to the baseline model, proving the effectiveness of our approach. The code is available at \url{https://github.com/sanmin0312/LabelDistill}

\end{abstract}

\label{sec:intro}

\section{Introduction}
\label{section:introduction}

3D object detection is an essential task in various applications, such as autonomous driving and robotics.
In recent years, camera-based methods \cite{wang2021fcos3d, park2021pseudo, wang2022detr3d, liu2022petr} have attracted extensive attention owing to their cost-effectiveness and rich semantic information that images can provide.
However, their current performance falls short when compared to LiDAR-based counterparts \cite{yin2021center, wang2021object, koh2023mgtanet}, primarily due to the absence of geometric and spatial information.

\begin{figure}[t]
\centering
\includegraphics[width=1\textwidth]{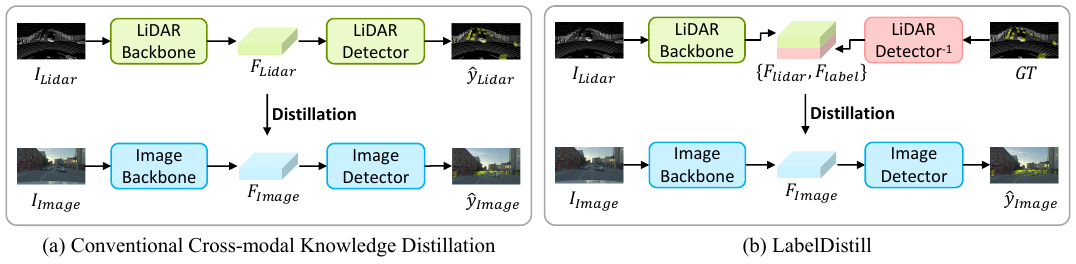}
% \vspace{-12pt}
\caption{
(a) \textbf{Conventional cross-modal knowledge distillation} trains an image detector to mimic the features of a well-trained LiDAR detector. 
It could be suboptimal as it directly transfers LiDAR features with inherent imperfections to the image feature. \hspace{3pt}
(b) \textbf{LabelDistill} enhances the image detector by incorporating ground truth labels into the feature representation. 
This approach aims to furnish the image detector with more accurate guidance, alleviating the intrinsic limitations of LiDAR point clouds. 
}
% \vspace{-10pt}

\label{figure:1}
\end{figure}

To bridge this performance gap between the camera and LiDAR detectors, knowledge distillation \cite{hinton2015distilling} emerges as a promising solution, following the success in various computer vision fields such as image classification \cite{yuan2020revisiting}, object detection \cite{yang2022focal} and segmentation \cite{liu2019structured}. 
Notably, LiDAR-guided cross-modal knowledge distillation methods \cite{klingner2023x3kd, li2022unifying, chenbevdistill, chongmonodistill, wang2023distillbev, hong2022cross, huang2023leveraging, jang2023stxd} hold great potential in the camera-based 3D object detection task.
These methods transfer learned information from LiDAR detectors to image detectors, leveraging precise spatial features from LiDAR without requiring LiDAR sensors during inference.

Despite the improvements observed in current LiDAR-guided cross-modal knowledge distillation methods, they are not without their limitations.
First, they tend to overlook the inherent imperfections of LiDAR point clouds, including aleatoric uncertainties in distant and occluded objects.
Such shortcomings make features from LiDAR detector imperfect for distillation.
Second, existing methods insufficiently handle complementary characteristics of LiDAR and camera.
While LiDAR provides precise spatial information, the camera offers abundant semantic information.
Therefore, indiscriminate distillation aiming to align all image features with LiDAR features may hinder the extraction of the full potential of image features.

To address these limitations, we present a novel cross-modal knowledge distillation approach tailored for camera-based 3D object detection. 
Our approach introduces a label distillation strategy that capitalizes on aleatoric uncertainty-free features derived from ground truth labels within the distillation process.
Unlike conventional label guidance approaches \cite{zhang2022lgd, hao2020labelenc}, which extract label features supervised by student features, our label distillation method focuses on extracting label features that can complement the limitations of LiDAR point clouds. 
This is achieved by leveraging the inverse function of a well-trained teacher's head, which can effectively map 3D bounding boxes into a teacher's feature space.
When combined with LiDAR distillation, our label distillation approach provides accurate and robust guidance to the image detector, enhancing its overall performance. 

Furthermore, we introduce a feature partitioning strategy in the distillation process to effectively transfer knowledge from the teacher modality while preserving the complementary features of the student modality, such as semantic information.
We separate student's features into several groups in the channel dimension, allocating some to the teacher while keeping others unaffected by the teacher.
This approach ensures that the student can learn informative features from the teacher without compromising its own unique characteristics.
In summary, the contributions of this paper are:
\begin{itemize}
\item We propose a novel label-guided cross-modal knowledge distillation, which effectively complements the imperfections of the LiDAR-based teacher model, leveraging the aleatoric uncertainty-free features.

\item We introduce a feature partitioning to effectively transfer knowledge from the teacher modality while preserving the distinctive information of the student modality.

\item Our approach achieves improved performance compared to prior state-of-the-art methods without incurring additional costs in the inference stage. Extensive experimentation confirms the effectiveness of our approach.

\end{itemize}

\section{Related Work}

\noindent \textbf{Camera-based 3D Object Detection.} 
Early approaches in camera-based 3D object detection \cite{brazil2019m3d, liu2020smoke, mousavian20173d, park2021pseudo, wang2021fcos3d} built upon the success of 2D detection methods \cite{tian2019fcos, zhou2019objects}.
These methods utilized perspective view features to directly estimate 3D information from 2D image inputs.
However, they faced the challenge of ill-posed depth estimation, stemming from information loss during the projection from 3D to 2D.
To mitigate such inaccurate depth estimation, several methods \cite{lu2021geometry, li2022diversity, qin2022monoground, wang2022probabilistic} have explored geometric information, while DD3D \cite{park2021pseudo} have incorporated depth pre-training using additional datasets \cite{guizilini20203d}.

Recent progress in the field have involved the adoption of of Bird's-Eye-View (BEV) feature representation through view transformation.
A line of works \cite{philion2020lift, reading2021categorical, huang2021bevdet, li2023bevdepth} has adopted forward view transformations by projecting perspective view features into BEV space using estimated depth distribution.
On the other hand, other works \cite{roddick2018orthographic, li2022bevformer, jiang2023polarformer, wang2022detr3d, chen2022polar, yang2023bevformer} have employed backward view transformation by incorporating attention mechanism \cite{vaswani2017attention} for correspondences between 3D and 2D space. Despite these advancements in camera-based 3D object detection showing promising performance, challenges persist in achieving accurate localization due to the inherent limitations of depth information.

\noindent \textbf{Knowledge Distillation for 3D Object Detection.}
Knowledge distillation is initially proposed for the model compression \cite{hinton2015distilling} by transferring the information from a large and cumbersome teacher model to a light and compact student model.
It has proven effective in various computer vision domains, such as classification \cite{park2019relational, xie2020self, yuan2020revisiting}, object detection \cite{dai2021general, cao2022pkd, yang2022focal}, and semantic segmentation \cite{liu2019structured, wang2020intra, tian2022adaptive}.
Recently, this strategy has been applied to the 3D object detection task \cite{zhang2023pointdistiller, cho2023itkd, zeng2023distilling, zhang2023qd}.

In autonomous driving applications, LiDAR-guided cross-modal knowledge distillation methods \cite{zhou2023unidistill, klingner2023x3kd, li2022unifying, chenbevdistill, chongmonodistill, wang2023distillbev, hong2022cross, huang2023leveraging, jang2023stxd} are gaining attention, which introduce a LiDAR detector as the teacher model to provide accurate and rich spatial information obtained from LiDAR point clouds to an image detector.
MonoDistill \cite{chongmonodistill} projects LiDAR points into the image plane to unify the representations, and BEVDistill \cite{chenbevdistill} introduces a sparse instance-wise distillation in addition to dense feature imitation.
On the other hand, X$^3$KD \cite{klingner2023x3kd} proposes cross-task knowledge distillation that transfers information from instance segmentation tasks.
Despite their promising results, these methods often overlook the imperfections in LiDAR data, leading to suboptimal distillation.
Additionally, domain discrepancies between LiDAR and camera modalities are insufficiently addressed. 

\noindent \textbf{Label Guidance.}
Several works across various tasks have integrated label guidance into their training schemes. One line of work \cite{newell2016stacked, mostajabi2018regularizing} employs labels for intermediate supervision, offering auxiliary guidance for regularization. 
Another line of work \cite{zhang2022lgd, liu2021label, huang2022label, hao2020labelenc} utilizes label input to enhance student features within a teacher-free distillation framework. 
However, these methods struggle to effectively extract useful features from labels as they typically employ simplistic autoencoders or rely on student features to train the label encoder, resulting in suboptimal label features.
In contrast, our approach involves embedding labels into the feature space of a LiDAR teacher model, thereby providing valuable label features that can complement teacher features.

\section{Method}

\begin{figure*}[t]
\centering
\includegraphics[width=0.95\textwidth]{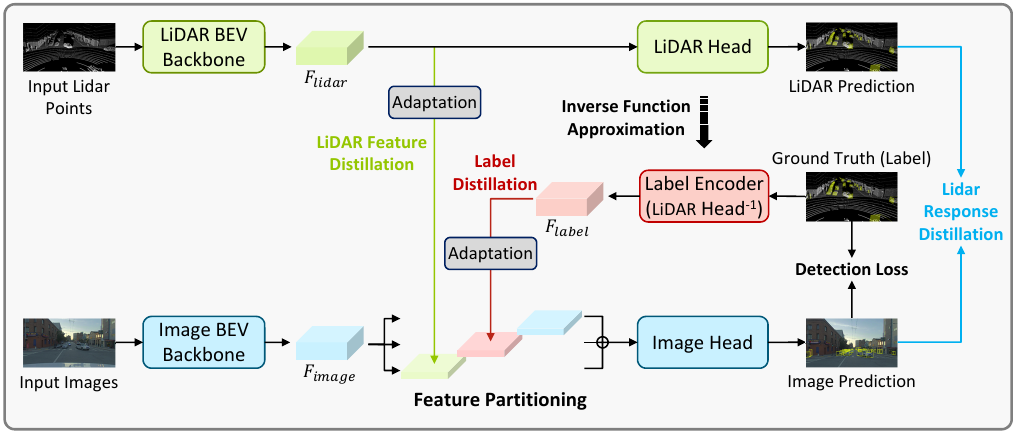}

% \vspace{-5pt}
\caption{
Overall architecture of the proposed method.
Our model is trained with two distillation strategies: LiDAR distillation and label distillation.
\textbf{LiDAR Distillation} transfers abundant spatial information to the image detector using feature-level and response-level distillation.
\textbf{Label Distillation} provides accurate and aleatoric uncertainty-free information based on the ground truth label to compensate the limitations of LiDAR point clouds.
In addition, \textbf{Feature Partitioning} separates the image features into three groups to preserve distinctive image features while learning from LiDAR and label features.
}
\vspace{-10pt}

\label{figure:2}
\end{figure*}

As illustrated in \cref{figure:2}, our proposed method consists of three pipelines: LiDAR, ground truth labels, and image.
The primary goal is to guide the image detector in learning accurate spatial information by employing label distillation in addition to LiDAR distillation, all while preserving its distinctive features.

\subsection{LiDAR Distillation}
The LiDAR distillation process follows the conventional knowledge distillation paradigm, utilizing a LiDAR detector as the teacher model. 
Our approach begins by extracting Bird's-Eye-View (BEV) features from both LiDAR point clouds and multi-view images, employing independent backbones for each modality.
We utilize two LiDAR distillation strategies: feature-level and response-level distillation.

\noindent \textbf{Feature-level Distillation.}
Feature-level distillation aims to transfer rich spatial and geometric information from LiDAR BEV features to the corresponding image BEV features.
These image BEV features are transformed from the perspective view using view transformation techniques \cite{philion2020lift, li2022bevformer}.
This distillation is facilitated through a loss function as follows:

\begin{equation}
\label{equation:lidar_feat}
    \mathcal{L}_{lidar}^{feat} = \frac{1}{N_{p}} \sum_{i}^{H} \sum_{j}^{W} \mathcal{M}_{ij}\{F_{ij}^{lidar}-\alpha(F_{ij}^{image})\}^{2},
\end{equation}

\noindent
where $H$ and $W$ represent the height and width of the BEV feature map.
$F_{ij}^{lidar}$ and $F_{ij}^{image}$ are the BEV features at location $(i,j)$ from the LiDAR and image, respectively. 
The mask $\mathcal{M}$ isolates the distillation process to object-specific regions, employing a foreground mask derived from the ground truth heatmap within the BEV space.
% Experiments for the foreground mask is demonstrated in \cref{table:mask}.
$N_p$ is the number of non-zero pixels in $\mathcal{M}$.
The adaptation module $\alpha$, consisting of convolutional layers, aligns the dimensionality of the image features with the teacher model's output.

\noindent \textbf{Response-level Distillation.}
In response-level distillation, the predictions from the LiDAR detector are used as an additional soft label, following \cite{hinton2015distilling}:

\begin{equation}
\label{equation:lidar_resp}
    \mathcal{L}_{lidar}^{resp} = \mathcal{L}_{cls}(c_{lidar}, c_{image}) + \mathcal{L}_{bbox}(b_{lidar}, b_{image}),
\end{equation}

\noindent
where $c$ and $b$ denote the class heatmap and bounding box predictions from LiDAR and image detector, respectively. 
We employ focal loss for the classification loss $\mathcal{L}_{cls}$ and L1 loss for the regression loss $\mathcal{L}_{bbox}$.
In this process, we utilize foreground masking based on ground truth heatmaps to prevent negative impacts from false positives.

\subsection{Label Distillation}
While LiDAR distillation provides essential spatial information to guide the image detector, the inherent limitations of LiDAR point clouds, such as ambiguity in distant or occluded objects due to sparsity \cite{zhang2021faraway} and susceptibility to adverse weather \cite{hahner2021fog, hahner2022lidar}, can potentially impact the quality of features used in the distillation process.
These imperfections tend to be neglected in existing studies since they were overshadowed by the superior detection performance of the LiDAR object detectors over camera detectors, thereby limiting the full potential of LiDAR-guided cross-modal knowledge distillation.
To overcome these limitations, we introduce label distillation as a complementary strategy alongside LiDAR distillation.
The label distillation leverages the ground truth labels.
The ground truth labels are generated by human annotators using multiple sensors with long sequential frames (\eg, the nuScenes dataset \cite{caesar2020nuscenes} leverages LiDAR, radar, and camera with 20 seconds of frames, including past and future timesteps).
As a result, these ground truth labels are ready to offer precise 3D object bounding boxes that are free from aleatoric uncertainty, providing the image detector with reliable guidance.

\noindent
\textbf{Approximating the Inverse Function of the Teacher's Head.}
A crucial step in mitigating the limitations of the teacher model is to adequately encode ground truth labels into the feature space. 
Previous efforts to utilize labels for guiding the training process have been explored in several works \cite{zhang2022lgd, huang2022label, hao2020labelenc}. 
However, these methods have often fallen short in extracting optimal features from label inputs, primarily due to a training process that forces label features to be similar to student features.
To address this challenge, we leverage the LiDAR detection head's capability that decode LiDAR features into 3D bounding box predictions:
% \vspace{-2pt}
\begin{equation}
\label{equation:lidar_head}
    \hat{y} = h(F_{lidar} ; \theta_{h} ),
\end{equation}

\noindent
where $F_{lidar}$ and $\hat{y}$ denote LiDAR features and the bounding box predictions, respectively.
$h(\cdot ; \theta_{h})$ represents the LiDAR detection head.

This process implies that the inverse function of the LiDAR detection head can map bounding box representations back into feature space. 
Accordingly, we aim to embed labels, which are 3D bounding boxes, into the feature space of the teacher model using this inverse function of the LiDAR detection head, as formalized in the following equation:

% \vspace{-5pt}
\begin{equation}
\label{equation:label_1}
    F_{label} = h^{-1}(y ;\theta_{h^{-1}}),
\end{equation}

\noindent
where $h^{-1}(\cdot ;\theta_{h^{-1}})$ represents the inverse function of the LiDAR detection head, acting as the label encoder.
In other words, $h^{-1}$ can output optimal label features given ground truth 3D bounding box inputs.

\begin{figure}[t]
\centering
\includegraphics[width=0.9\textwidth]{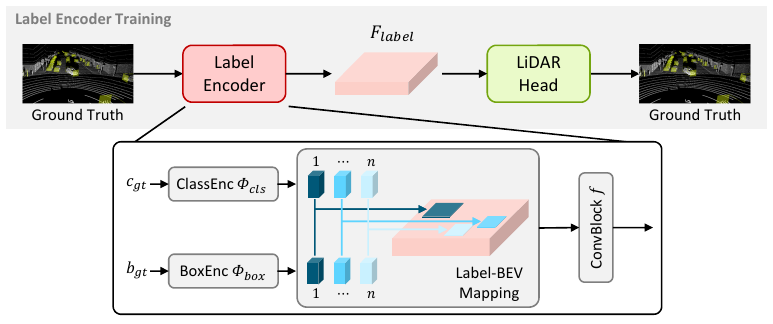}
% \vspace{-5pt}

\caption{
Architecture of the label encoder.
The label encoder is designed to approximate the inverse function of the pretrained lidar detection head.
The label encoder first encodes class and bounding box information and then, the mapping function transforms encoded label features into BEV space by filling the object's bounding box area with label features.
Finally, the convolutional block encodes BEV label features.
}

\vspace{-5pt}

\label{figure:3}

\end{figure}

However, calculating this inverse function is impractical due to the high non-linearity of neural networks.
Inspired by \cite{hao2020labelenc} and \cite{mostajabi2018regularizing}, we utilize an autoencoder framework to approximate the inverse function of the LiDAR detection head.
Within this framework, the label encoder assumes the role of the encoder, and the pre-trained LiDAR detection head functions as the decoder, as described in \cref{figure:3}.
The training objective for the label encoder is formulated as:

\begin{equation}
\label{equation:label_training}
% \theta_{g}^{*} = \underset{\theta_{g}}{\arg\min} \mathbb{E}_{(I_{lidar},y) \sim \mathcal{D}} \; \; \mathcal{L}_{det}(h(g(y;\theta_{g});\theta_{h^{*}}),y)
\theta_{g}^{*} = \underset{\theta_{g}}{\arg\min} \mathbb{E}_{(I,y) \sim \mathcal{D}} \;\; \mathcal{L}_{det}\Bigl(h\bigl(g(y;\theta_{g});\theta_{h}^{*}\bigr),y\Bigr),
\end{equation}

\noindent
where $h(\cdot ;\theta_{h}^{*})$ represents the pretrained LiDAR detection head, and $g(\cdot ;\theta_{g})$ represents the label encoder designed to approximate $h^{-1}(\cdot ;\theta_{h}^{*})$.
$(I,y)$ denotes a pair of LiDAR point cloud and ground truth label, $\mathcal{D}$ is the distribution of the dataset, and $\mathcal{L}_{det}(\cdot, \cdot)$ represents the detection loss function for the classification and bounding box regression.
In this manner, label distillation effectively mitigates imperfections of LiDAR point clouds by aligning the aleatoric uncertainty-free label features to the teacher's feature space.
Our approach differs from conventional autoencoders by setting the decoder as the pretrained LiDAR detection head and focusing the training on the encoder (the label encoder), whereas a standard autoencoder would train both components from scratch.

\noindent
\textbf{Label Encoder.}
As depicted in \cref{figure:3}, we have adopted a simple design for the label encoder due to the compact and noise-free nature of ground truth labels.
The label encoder handles both class and bounding box information, employing a straightforward yet efficient structure for label encoding.
The label encoder is defined as follows:

\begin{equation}
\label{equation:label_model}
g(y ;\theta_{g}) = f\Bigl(q\bigl(\Phi_{cls}(c_{gt}) + \Phi_{box}(b_{gt})\bigr)\Bigr),
\end{equation}

\noindent
where $c_{gt} \in \mathbb{R}^{n \times m}$ represents the ground truth class information of $m$ classes for $n$ objects in the scene, while $b_{gt} \in \mathbb{R}^{n \times z}$ is the ground truth bounding box information of $z$ attributes such as 3D location, size, orientation and velocity.
$\Phi_{cls}$ and $\Phi_{box}$ are MLP layers to embed class and bounding box information.
The embedded class and bounding box vectors are placed in foreground on the BEV space using the mapping function $q(\cdot)$ after summation.
We fill each BEV grid occupied by the bounding box of an object, with duplicated label feature vectors, to generate the label BEV feature.
Subsequently, the function $f$, which is the convolutional block including convolutional layers, normalization, and activations, is employed to refine the feature maps into the final label feature $F_{label}$.
Note that the label encoder is pretrained before the distillation process.

The implementation of this label encoder has proven to be highly effective in approximating the inverse function of the LiDAR detection head, achieving a 94\% mean Average Precision (mAP) when combined with the LiDAR detection head, as illustrated in \cref{table:label_encoder}.
This result also demonstrates that the encoded label features preserve useful information to reconstruct 3D bounding boxes while mapping to the teacher's feature space.

%==========================================================================================================================================
%Label encoder
%==========================================================================================================================================

\begin{table}[t]
\begin{center}
\caption{
Evaluation of the autoencoder consists of the label encoder and LiDAR detection head on nuScenes validation set.
} 
\label{table:label_encoder}
\setlength{\tabcolsep}{5pt}
\resizebox{0.8\textwidth}{!}
{
\renewcommand{\arraystretch}{1.2}
\begin{tabular}{c|cc|ccc}

\hline
 & \textbf{mAP} $\uparrow$  & \textbf{NDS} $\uparrow$  & \textbf{mATE} $\downarrow$ & \textbf{mAOE} $\downarrow$ & \textbf{mAVE} $\downarrow$ \\
\hline
\hline

\begin{tabular}[c]{@{}l@{}}\textbf{Label Encoder} \\ \textbf{+ LiDAR Head}\end{tabular} & 94.14 & 90.25 & 0.192 & 0.048 & 0.128 \\
\hline

\end{tabular}
}
\end{center}

\vspace{-17pt}
\end{table}
%==========================================================================================================================================

\subsection{Feature Partitioning}
LiDAR point clouds are a rich source of precise spatial and geometric data, while images offer dense semantic details.
These modalities are inherently complementary. 
However, conventional cross-modal distillation approaches that attempt to train all image feature channels to mimic LiDAR features may not fully harness the potential of images.
Additionally, our approach utilizes multiple teachers, including LiDAR and label, which can potentially result in contrasting supervision due to the disparate nature of features.

To address these challenges, we introduce a straightforward yet effective strategy: feature partitioning. 
This strategy aims to preserve distinctive image features while simultaneously learning from the LiDAR and label features.
We partition the image feature $F_{\text{image}} \in \mathbb{R}^{H \times W \times C}$ into three distinct groups along the channel dimension: ${F_{image}^{image}, F_{image}^{lidar}, F_{image}^{label}}$.
Each feature group consists of a subset of the image features with a combined total of $C$ channels.
The group $F_{image}^{lidar}$ is designed to focus on learning essential LiDAR features, leveraging the spatial and geometric details provided by the LiDAR data.
Meanwhile, the group $F_{image}^{label}$ is dedicated to learning label-related features.
In contrast, the group $F_{image}^{image}$ remains unaffected by the influence of the teacher models.
This group is exclusively trained using the detection loss function.
By remaining uninfluenced by the teacher models, this group retains the inherent semantic features found in the image data, ensuring that the richness and depth of the semantic information remain intact throughout the training process.

\subsection{Training}
Our model undergoes a two-step training process. 
In the first step, we train the label encoder to approximate the inverse function of the pretrained LiDAR detection head. 
During this step, the label encoder is trained using a conventional detection loss, including classification and bounding box regression losses.
In the second step, we train the image detector with the pretrained label encoder and LiDAR detector. 
This step involves training our model with a loss function comprising four terms: LiDAR feature loss, LiDAR response loss, label feature loss, and detection loss, which is formulated as:

\begin{equation}
\label{equation:loss}
    \mathcal{L}= \mathcal{L}_{det} + \lambda_{1}\mathcal{L}_{lidar}^{feat} + \lambda_{2}\mathcal{L}_{label}^{feat}+ \lambda_{3}\mathcal{L}_{lidar}^{resp},
\end{equation}

\noindent
where $\lambda_{1,2,3}$ are balancing weight term.
We adopt the same loss function as presented in \cite{li2023bevdepth} for the detection loss.
It consists of classification loss, bounding box regression loss, and depth loss.
Meanwhile, the label feature loss employs the Mean Squared Error (MSE) loss with foreground masking, similar to the LiDAR feature loss in \cref{equation:lidar_feat}.
It is worth to note that our distillation strategies do not introduce any additional computational burden during the inference stage.

\section{Experiments}

\subsection{Experimental Setup}

\noindent \textbf{Dataset and Metrics.}
We train and evaluate our approach on the nuScenes dataset \cite{caesar2020nuscenes}, which is the large-scale autonomous driving benchmark.
It consists of 1000 videos of around 20 seconds with annotations at 2Hz, including 3D bounding boxes of 10 classes.
We follow the official evaluation metrics to evaluate 3D object detection performance, including mean Average Precision (mAP) and nuScenes Detection Score (NDS). 
We also report other metrics such as mean Average Translation Error (mATE), mean Average Scale Error (mASE), mean Average Orientation Error (mAOE), mean Average Velocity Error (mAVE), and mean Average Attribute Error (mAAE).

\noindent \textbf{Teacher and Student Model.}
For teacher model, we adopt pretrained CenterPoint \cite{yin2021center} with a voxel size of (0.1m, 0.1m, 0.2m).
For student model, we employ BEVDepth \cite{li2023bevdepth}.
Unless otherwise specified, ResNet50 pretrained with ImageNet is adopted as image backbone, and the input image is resized to 256 $\times$ 704. 
% The student model with ResNet101 adopts input image size of 512 $\times$ 1408.
We follow the image and BEV data augmentation strategies in \cite{li2023bevdepth}.
We use four previous frames for the experiments of \cref{table:valset} and \cref{table:distillation} while one previous frame is adopted for ablation studies.
% Both teacher and student use BEV feature size of 128 $\times$ 128.

\noindent \textbf{Implementation Details.}
The label encoder is trained for 12 epochs with the learning rate of 1e-3 while 24 epochs and learning rate of 4e-4 is employed for training the image detector. 
We adopt AdamW optimizer \cite{loshchilov2017adamw} without CBGS \cite{zhu2019cbgs}. 
A batch size of 16 on 4 NVIDIA 3090Ti GPUs is used for both training of label encoder and distillation of student model. 

%==========================================================================================================================================
% nuscenes val set
%==========================================================================================================================================

\begin{table}[t]

\begin{center}
\caption{Comparison on the nuScenes dataset. $\dagger$: methods with CBGS. $^{*}$: reproduced with the same setting as our model for a fair comparison.} 
\label{table:valset}
\vspace{-10pt}
\resizebox{1.0\textwidth}{!}{
\setlength{\tabcolsep}{1.2pt}
\renewcommand{\arraystretch}{1.3}

\begin{tabular}{c|l|cc|cc|ccccc}

\hline
\rowcolor{white}
\textbf{Set} & \textbf{Method} & \textbf{Backbone} & \textbf{Size} & \textbf{mAP}  & \textbf{NDS} & \textbf{mATE} & \textbf{mASE} & \textbf{mAOE} & \textbf{mAVE} & \textbf{mAAE}\\ 
\hline
\hline

{\multirow{11}{*}{\rotatebox[origin=c]{90}{Validation}}} & 
 BEVDet4D \cite{huang2022bevdet4d}    & ResNet50 & 256$\times$704    & 32.3 & 45.3 & 0.674 & 0.272 & 0.503 & 0.429 & 0.208 \\
& BEVDepth \cite{li2023bevdepth}       & ResNet50 & 256$\times$704    & 33.3 & 44.1 & 0.683 & 0.276 & 0.545 & 0.526 & 0.226 \\
& BEVStereo \cite{li2023bevstereo}     & ResNet50 & 256$\times$704    & 34.4 & 44.9 & 0.659 & 0.276 & 0.579 & 0.503 & 0.216 \\ 
& VEDet$^\dagger$ \cite{chen2023viewpoint} & ResNet50 & 384$\times$1056    & 34.7 & 44.3 & 0.726 & 0.282 & 0.542 & 0.555 & 0.198 \\
& PETR v2 \cite{liu2023petrv2}          & ResNet50 & 256$\times$704    & 34.9 & 45.6 & 0.700 & 0.275 & 0.580 & 0.437 & \textbf{0.187} \\ 
& FB-BEV$^\dagger$ \cite{li2023fb}               & ResNet50 & 256$\times$704    & 35.0 & 47.9 & 0.642 & 0.275 & 0.459 & 0.391 & 0.193 \\ 
& AeDet$^\dagger$ \cite{feng2023aedet}           & ResNet50 & 256$\times$704    & 35.8 & 47.3 & 0.655 & 0.273 & 0.493 & 0.427 & 0.216 \\ 
& P2D \cite{kim2023predict}            & ResNet50 & 256$\times$704    & 37.4 & 48.6 & 0.631 & 0.272 & 0.508 & 0.384 & 0.212 \\ 
& BEVFormer v2$^\dagger$\cite{yang2023bevformer} & ResNet50 & 640$\times$1600    & 38.8 & 49.8 & 0.679 & 0.276 & 0.417 & 0.403 & 0.189 \\ 
& SOLOFusion \cite{park2022time} & ResNet50 & 256$\times$704    & 40.6 & 49.7 & 0.609 & 0.284 & 0.650 & \textbf{0.315} & 0.204 \\ 

& \cellcolor{Gray}LabelDistill   & \cellcolor{Gray}ResNet50 & \cellcolor{Gray}256$\times$704    & \cellcolor{Gray}\textbf{41.9} & \cellcolor{Gray}\textbf{52.8} & \cellcolor{Gray}\textbf{0.582} & \cellcolor{Gray}\textbf{0.258} & \cellcolor{Gray}\textbf{0.413} & \cellcolor{Gray}0.346 & \cellcolor{Gray}0.220 \\ 

\hline
\rowcolor{white}
{\multirow{8}{*}{\rotatebox[origin=c]{90}{Validation}}} & DETR3D$^\dagger$  \cite{wang2022detr3d} & ResNet101 & 900$\times$1600    & 34.9 & 43.4 & 0.716 & 0.268 & 0.379 & 0.842 & 0.200 \\
& BEVDepth \cite{li2023bevdepth}      & ResNet101 & 512$\times$1408    & 40.6 & 49.0 & 0.626 & 0.278 & 0.513 & 0.489 & 0.226 \\
& BEVFormer \cite{li2022bevformer}        & ResNet101 & 900$\times$1600    & 41.6 & 51.7 & 0.673 & 0.274 & 0.372 & 0.394 & 0.198 \\
& VEDet$^\dagger$ \cite{chen2023viewpoint} & ResNet101 & 512$\times$1408    & 43.2 & 52.0 & 0.638 & 0.275 & 0.362 & 0.498 & 0.191 \\
& PolarFormer \cite{jiang2023polarformer} & ResNet101 & 900$\times$1600    & 43.2 & 52.8 & 0.648 & 0.270 & 0.348 & 0.409 & 0.201 \\
& P2D \cite{kim2023predict}               & ResNet101 & 512$\times$1408    & 43.3 & 52.8 & 0.619 & 0.265 & 0.432 & 0.364 & 0.211 \\
& Sparse4D \cite{lin2022sparse4d}         & ResNet101 & 900$\times$1600    & 43.6 & 54.1 & 0.633 & 0.279 & 0.363 & \textbf{0.317} & \textbf{0.177} \\  

& \cellcolor{Gray}LabelDistill  & \cellcolor{Gray}ResNet101 & \cellcolor{Gray}512$\times$1408    & \cellcolor{Gray}\textbf{45.1} & \cellcolor{Gray}\textbf{55.3} & \cellcolor{Gray}\textbf{0.579} & \cellcolor{Gray}\textbf{0.252} & \cellcolor{Gray}\textbf{0.331} & \cellcolor{Gray}0.357 & \cellcolor{Gray}0.207 \\
\hline

{\multirow{2}{*}{\rotatebox[origin=c]{90}{Test}}} & BEVDepth$^{*}$ \cite{li2023bevdepth} & ConvNeXt-B & 900$\times$1600 & 47.5 & 56.1 & 0.474 & 0.259 & 0.463 & 0.432 & \textbf{0.134}\\
& \cellcolor{Gray}LabelDistill & \cellcolor{Gray}ConvNeXt-B & \cellcolor{Gray}900$\times$1600 & \cellcolor{Gray}\textbf{52.6} & \cellcolor{Gray}\textbf{61.0} & \cellcolor{Gray}\textbf{0.443} & \cellcolor{Gray}\textbf{0.241} & \cellcolor{Gray}\textbf{0.339} & \cellcolor{Gray}\textbf{0.370} & \cellcolor{Gray}0.136\\
\hline

\end{tabular}
}
\end{center}
\vspace{-20pt}

\end{table}
%==========================================================================================================================================

% %Comparison on LiDAR Disillation
% %==========================================================================================================================================

\begin{table}[t]
\begin{center}

\caption{
Comparison to other LiDAR-guided cross-modal knowledge distillation strategies. 
$\dagger$: methods with CBGS.
} 
\label{table:distillation}
% \vspace{-12pt}
\resizebox{0.93\textwidth}{!}
{
\setlength{\tabcolsep}{3pt}
\renewcommand{\arraystretch}{1.2}
\begin{tabular}{l|ccc|cc}
\rowcolor{white}

\hline
\textbf{Model} & \textbf{Baseline} & \textbf{Image Size} & \textbf{Backbone} & \textbf{mAP} ($\Delta$) &\textbf{NDS} ($\Delta$)\\ 
\hline
\hline

UniDistill \cite{zhou2023unidistill}& BEVDet & 704$\times$256 & ResNet50 & 29.6 (3.2) & 39.3 (3.2)\\ % 3.2/3.2
BEVDistill \cite{chenbevdistill}& BEVDepth & 704$\times$256 & ResNet50 & 33.0 (1.3) & 45.2 (1.2)\\ % 1.3/1.2
% UVTR & ResNet50 &  & 36.2 & 47.2\\
TiG-BEV \cite{huang2022tig} & BEVDepth & 704$\times$256 & ResNet50 & 36.6 (3.7) & 46.1 (3.0) \\ % 3.7 / 3.0
BEVSimDet \cite{zhao2023bevsimdet}& BEVFusion-C & 704$\times$256 & ResNet50 & 37.3 (1.7) & 43.8 (2.6) \\ % 1.7 / 2.6
X$^{3}$KD$^\dagger$ \cite{klingner2023x3kd} & BEVDepth & 704$\times$256 & ResNet50 & 39.0 (3.1) & 50.5 (3.3)\\ % 3.1/3.3
DistillBEV$^\dagger$ \cite{wang2023distillbev}& BEVDepth & 704$\times$256 & ResNet50 & 40.3 (3.9) & 51.0 (2.6)\\ % 3.9/2.6

\cellcolor{Gray}LabelDistill  & \cellcolor{Gray}BEVDepth & \cellcolor{Gray}704$\times$256  & \cellcolor{Gray} ResNet50 & \cellcolor{Gray}\textbf{41.9 (5.1)} & \cellcolor{Gray}\textbf{52.8 (4.5)}\\

\hline
UVTR \cite{li2022unifying} & - & 1600$\times$900 & ResNet101 & 39.2 (1.3) & 48.8 (0.5)\\ % 1.3/0.5
BEVDistill$^\dagger$ \cite{chenbevdistill} & BEVFormer & 1600$\times$900 & ResNet101 & 41.7 (1.2) & 52.4 (1.8)\\ % 1.2/1.8
TiG-BEV \cite{huang2022tig} & BEVDepth & 1408$\times$512 & ResNet101 & 43.0 \textbf{(2.4)} & 51.4 (2.3) \\ % 2.4/2.3
DistillBEV$^\dagger$ \cite{wang2023distillbev} & BEVDepth & 1408$\times$512 & ResNet101 & 45.0 (2.3) & 54.7 (3.1)\\ % 2.3/3.1

\cellcolor{Gray}LabelDistill  & \cellcolor{Gray}BEVDepth & \cellcolor{Gray}1408$\times$512  & \cellcolor{Gray} ResNet101 & \cellcolor{Gray}\textbf{45.1 (2.4)} & \cellcolor{Gray}\textbf{55.3 (3.7)}\\

\hline
\end{tabular}
}
\end{center}
\vspace{-10pt}
\end{table}

%%==========================================================================================================================================

\begin{table}[h!]
\begin{center}
\caption{
Ablation study on the proposed method.
LiDAR, Label, and Partition represent LiDAR distillation, label distillation, and feature partitioning, respectively.
} 
\label{table:ablation}
% \vspace{-10pt}
\resizebox{0.8\textwidth}{!}
{
\setlength{\tabcolsep}{2pt}
\renewcommand{\arraystretch}{1.2}
\begin{tabular}{c|ccc|cc|cc}

\hline
& \textbf{LiDAR} & \textbf{Label} & \textbf{Partition} & \textbf{mAP} $\uparrow$&\textbf{NDS} $\uparrow$& \textbf{mATE} $\downarrow$ & \textbf{mASE} $\downarrow$\\ 
\hline
\hline
(a) &            &             &             & 33.6 & 44.8 & 0.694 & 0.273\\ 
(b) &\checkmark &             &             & 35.4 & 48.6 & 0.648 & 0.262\\ 
(c) & \checkmark & \checkmark  &             & 37.0 & 49.5 & 0.663 & 0.258\\ 
(d) & \checkmark & \checkmark  & \checkmark  & \textbf{37.9} & \textbf{50.1} & \textbf{0.641} & \textbf{0.256}\\ 
\hline

\end{tabular}
}
\end{center}
\vspace{-17pt}
\end{table}
% 

%==========================================================================================================================================

\subsection{Main Results}
We start our analysis by comparing our model with existing camera-based 3D object detection models on the nuScenes validation set.
As reported in \cref{table:valset}, our model achieves a significant improvement of 8.6\%p in mAP and 8.7\%p in NDS compared to the baseline model, BEVDepth, in the ResNet50 settings.
Notably, these improvements remain consistent with a 4.5\%p and 6.3\%p boost in mAP and NDS, respectively, even in the  ResNet101.
Furthermore, it also demonstrates superior performance compared to other state-of-the-art approaches.
It is noteworthy that our model attains these results without resorting to CBGS~\cite{zhu2019cbgs}, a data augmentation strategy that effectively extends a single epoch into 4.5 epochs.
 
In addition, we perform a comparative analysis of our model with other LiDAR-guided cross-modal knowledge distillation methods, as shown in Table \cref{table:distillation}.
For a fair comparison, we present performance gain from baselines ($\Delta$) for both mAP and NDS.
This metric allows for a simple and equitable comparison as each model shares the same experimental settings with its baseline.
As shown in \cref{table:distillation}, our approach achieves superior performance compared to these models.

In the case of the test set, we trained BEVDepth \cite{li2023bevdepth} with the same settings as our model to ensure fair comparison. As a results, our LabelDistill achieves improvement of 5.1\%p and 4.9\%p for mAP and NDS, respectively.

\subsection{Ablation Study}

We performed a series of comprehensive ablation studies to evaluate the contribution of individual components and the impact of different hyperparameters within our model. 
These studies were conducted on the nuScenes validation set, with results detailed in \cref{table:ablation} through \cref{table:distance}.

\begin{figure*}[t]
\centering
\includegraphics[width=0.95\textwidth]{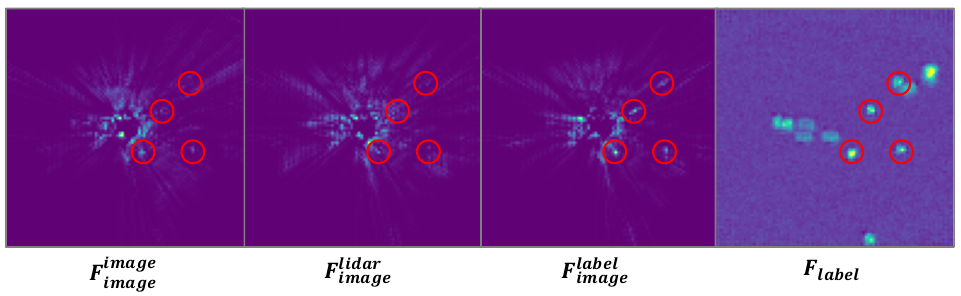}

% \vspace{-5pt}
\caption{
Illustration of BEV feature maps in the inference stage. $F_{image}^{image}$ is undistilled image feature, $F_{image}^{lidar}$ is lidar-distilled image feature, and $F_{image}^{label}$, label-distilled image feature, and $F_{label}$ denotes label feature from the label encoder.
}
% \vspace{-5pt}

\label{figure:feature}
\end{figure*}

\begin{table}[t]
\begin{center}
\caption{Experiments on different channel ratio for the feature partitioning.} 
\label{table:channel}
\resizebox{0.72\textwidth}{!}
{
\setlength{\tabcolsep}{3pt}
\renewcommand{\arraystretch}{1.2}
\begin{tabular}{ccc|cc|cc}
\hline
\multicolumn{3}{c|}{\textbf{Channel Ratio}} & \multirow{2}{*}{\textbf{mAP} $\uparrow$} & \multirow{2}{*}{\textbf{NDS} $\uparrow$} & \multirow{2}{*}{\textbf{mATE} $\downarrow$} & \multirow{2}{*}{\textbf{mASE} $\downarrow$} \\ \cline{1-3}
$F_{lidar}^{image}$ & $F_{label}^{image}$ & $F_{image}^{image}$ &  &  &  &  \\  
\hline
\hline

1  & 3  & 2  & 36.6 & 48.8 & 0.655 & 0.260 \\ 
3  & 1  & 2  & 37.1 & 49.4 & 0.646 & 0.258 \\ 
2  & 2  & 2  & \textbf{37.6} & \textbf{49.6} & \textbf{0.643} & \textbf{0.256} \\ 

\hline

\end{tabular}
}
\end{center}
\vspace{-20pt}
\end{table}

%==========================================================================================================================================

%==========================================================================================================================================
%inverse function approximation
%==========================================================================================================================================

\begin{table}[t]
\begin{center}
\caption{
Evaluation on the effectiveness of the inverse function approximation.
AutoEncoder trains the label encoder and the detection head from the scratch.
} 
\label{table:inverse}
\setlength{\tabcolsep}{4pt}
\resizebox{0.9\textwidth}{!}
{
\renewcommand{\arraystretch}{1.2}
\begin{tabular}{c|cc|ccc}
\hline
\textbf{Label Encoder Training} & \textbf{mAP} $\uparrow$ & \textbf{NDS} $\uparrow$ & \textbf{mATE} $\downarrow$ & \textbf{mASE} $\downarrow$ & \textbf{mAOE} $\downarrow$ \\
\hline
AutoEncoder & 34.9 & 46.7 & 0.656 & 0.270 & 0.476 \\
LabelEnc \cite{hao2020labelenc} & 34.8 & 46.8 & 0.658 & 0.267 & 0.479 \\
Inverse Function Approximation & \textbf{36.8} & \textbf{48.1} & \textbf{0.646} & \textbf{0.263} & \textbf{0.474} \\
\hline

\end{tabular}
}
\end{center}
\vspace{-7pt}
\end{table}
%==========================================================================================================================================

%==========================================================================================================================================
% label encoder ablation
%==========================================================================================================================================
\begin{table}[t]
\begin{center}
\caption{Experiments of the label encoder's impact on the student model.
Performance of the label encoder denotes AutoEncoder's performance, which consists of the label encoder and the LiDAR detection head.} 
\label{table:label_encoder_ab}
\resizebox{0.8\textwidth}{!}
{
\setlength{\tabcolsep}{5pt}
\renewcommand{\arraystretch}{1.2}
\begin{tabular}{cc|cc|ccc}

\hline
\multicolumn{2}{c|}{\begin{tabular}[c]{@{}l@{}}\textbf{Label Encoder} \\ \textbf{+ LiDAR Head}\end{tabular}} & \multicolumn{5}{c}{\textbf{Student Model}} \\ 
\hline
\textbf{mAP} $\uparrow$ & \textbf{NDS} $\uparrow$ & \textbf{mAP} $\uparrow$ & \multicolumn{1}{c|}{\textbf{NDS} $\uparrow$} & \textbf{mATE} $\downarrow$ & \textbf{mASE} $\downarrow$ & \textbf{mAOE} $\downarrow$ \\
\hline
\hline
50.2 & 42.9 & 34.0 & \multicolumn{1}{c|}{45.3} & 0.678 & 0.273  & 0.587 \\ %version7/5.pth
71.9 & 54.7 & 34.6 & \multicolumn{1}{c|}{45.6} & 0.673 & 0.274  & 0.583 \\ %version7/7.pth
94.1 & 90.3 & \textbf{36.8} & \multicolumn{1}{c|}{\textbf{48.1}} & \textbf{0.660} & \textbf{0.264}  & \textbf{0.470}\\ %version4/11.pth
\hline

\end{tabular}
}
\end{center}
\vspace{-7pt}
\end{table}

%==========================================================================================================================================

%==========================================================================================================================================
% distance ablation
%==========================================================================================================================================

\begin{table}[h!]
\begin{center}
\caption{Performance along the object distance. } 
\label{table:distance}

\resizebox{0.7\textwidth}{!}
{
\setlength{\tabcolsep}{5pt}
\renewcommand{\arraystretch}{1.2}
\begin{tabular}{c|cc|ccc}

\hline
\textbf{Distance} & \textbf{LiDAR} & \textbf{Label}  & \textbf{mATE} $\downarrow$ & \textbf{mASE} $\downarrow$ & \textbf{mAOE} $\downarrow$ \\
\hline
\hline

\multirow{2}{*}{$\leq$ 30m} & \checkmark &            & 0.592 & 0.261 & 0.397 \\
                            & \checkmark & \checkmark & 0.582 & 0.253 & 0.380 \\
\hline

\multirow{2}{*}{30m $\leq$} & \checkmark &            & 1.043 & 0.342 & 0.534 \\
                            & \checkmark & \checkmark & 1.012 & 0.270 & 0.531 \\

\hline

\end{tabular}
}
\end{center}
\vspace{-15pt}
\end{table}

%==========================================================================================================================================

% \input{tables/table_mask}

\noindent \textbf{Label Distillation.}
The ablation comparison presented in \cref{table:ablation} provides a analysis of the effectiveness of each strategy within our proposed model. 
LiDAR distillation (b) demonstrates improvement in performance compared to the baseline model. 
However, the integration of label distillation alongside LiDAR distillation (c) yields further enhancement, highlighting the capacity of label distillation to address the limitations of the LiDAR teacher model.
Moreover, we offer visual insights into the effectiveness of label distillation through the visualization of Bird's Eye View (BEV) features, as illustrated in \cref{figure:feature}. 
As depicted in \cref{figure:feature}, the label-distilled student feature ($F_{image}^{label}$) exhibits clear activation, whereas the lidar-distilled student feature ($F_{image}^{lidar}$) displays either blurry or negligible activation for occluded or distant objects. 
This observation underscores the superior capability of label distillation in capturing crucial information for challenging scenarios where LiDAR-based features may fall short.

\noindent \textbf{Feature Partitioning.}
The significance of feature partitioning is underscored by the comparison between (c) and (d) as depicted in \cref{table:ablation}. 
This comparison highlights the advantages conferred by feature partitioning within the distillation process, reaffirming its role in preserving the distinctive image features.

\noindent \textbf{Channel Ratio.}
We explore the impact of channel ratios on the feature partitioning strategy, maintaining a constant channel ratio for image features while varying the ratios for LiDAR and label features. 
We adopt 300 as the total channels, and as indicated in \cref{table:channel}, the most balanced performance is achieved when the channel ratios for all three features are identical.

\noindent \textbf{Inverse Function Approximation.}
To evaluate the effectiveness of training the label encoder by approximating the inverse function of the LiDAR detection head, we compared it with other label guidance methods, as shown in \cref{table:inverse}. 
AutoEncoder represents a simplistic approach where both an encoder and decoder are trained from scratch using labels as both inputs and targets. 
Similarly, LabelEnc \cite{hao2020labelenc} employs an AutoEncoder but integrates an additional encoding strategy that relies on the student feature during the label feature training process.
In contrast, our method leverages the inverse function of the teacher's head to effectively embed label features into the feature space. 
As demonstrated in \cref{table:inverse}, our approach outperforms other label guidance methods. 
These results underscore the effectiveness of employing the inverse function approximation of the teacher head, which ensures accurate and noise-free features are provided to the student model during the distillation process.

\noindent \textbf{Impact of Label Encoder Performance.}
We examined the influence of the label encoder's performance on the distillation process.
By deliberately reducing the label encoder's detection capabilities during the label encoder training, we observed a positive correlation between the label encoder's performance and that of the distilled student model, as shown in \cref{table:label_encoder_ab}.
This observation underscores the significance of an accurate inverse function approximation of the teacher detection head in providing precise label features in our distillation strategy.

\noindent \textbf{Distant Objects.}
An evaluation based on object distance was performed to further explore label distillation's impact, with findings shown in \cref{table:distance}. 
The results confirm an overall performance enhancement in models using label distillation.
Notably, the size estimation accuracy (mASE) for distant objects (over 30m) is substantially improved when employing label distillation as opposed to solely LiDAR distillation. 
This improvement can be attributed to the mitigating effect of label distillation on LiDAR sparsity.
The sparsity inherent in LiDAR often results in limited points being reflected from distant objects, making size estimation challenging. 
However, the label distillation resolves this challenge by providing accurate and reliable information, thereby alleviating the impact of sparsity, particularly for distant objects.

\begin{figure*}[t]
\centering
\includegraphics[width=0.95\textwidth]{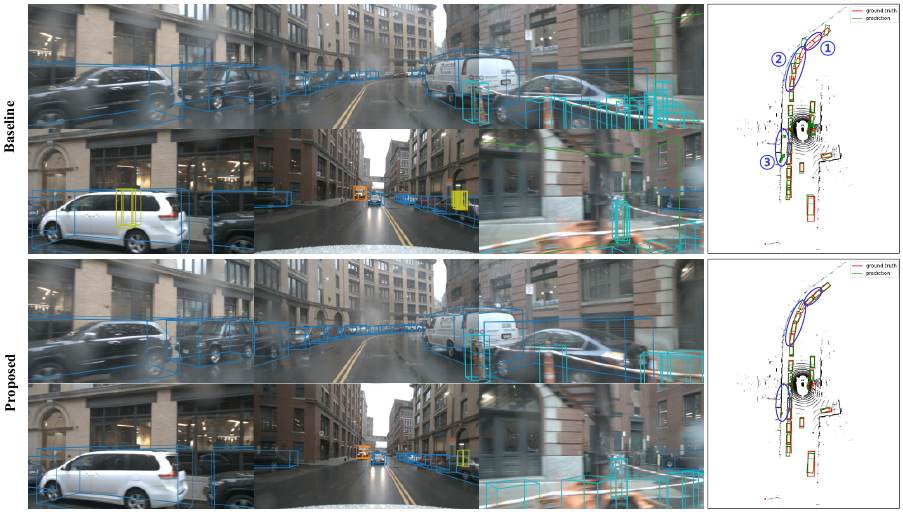}
% \vspace{-5pt}

\caption{
Comparison of the baseline (BEVDepth) and our approach.
The blue circles in the BEV view highlight cases that demonstrate the advantages of our approach, including: 1) higher recall, 2) more accurate localization, and 3) fewer false positives.
}
\vspace{-10pt}

\label{figure:4}
\end{figure*}

\subsection{Qualitative Results}

In \cref{figure:4}, we visualize a sample case to compare our approach to the baseline model.
As indicated with blue circles, LabelDistill demonstrates several key advantages: 
1) It achieves higher recall by successfully detecting objects that the baseline model often misses.
2) The accuracy of object localization is notably enhanced. 
LabelDistill accurately detects the location of objects, whereas the baseline model tends to yield imprecise results.
3) It effectively reduces false positives. 
LabelDistill reduces unnecessary and redundant bounding boxes while the baseline generates multiple redundant bounding boxes along the depth direction due to its inaccurate depth estimation ability.
These advantages make LabelDistill a promising solution for enhancing camera-based 3D object detection in real-world applications.

\section{Conclusion}

In this paper, we have presented a novel approach for cross-modal knowledge distillation aimed at effectively transferring knowledge from a LiDAR detector to an image detector. 
Our method, LabelDistill, addresses the inherent imperfections of LiDAR detectors by leveraging precise ground truth labels to provide accurate and aleatoric uncertainty-free features. 
Additionally, we have introduced a feature partitioning strategy designed to preserve distinctive image features while simultaneously facilitating the learning of accurate spatial information from the teacher model. 
Our extensive experiments demonstrate the effectiveness of the proposed methods.
However, the performance of the proposed method still lag behind compared to those of LiDAR detector.

However, it is important to note that the performance of the proposed method still lags behind compared to those of LiDAR detectors. 
Furthermore, the effectiveness of our method is dependent on the quality of the ground truth labels. 
If the ground truth labels in the dataset exhibit low reliability, the performance of the proposed method may be degraded.\\

\noindent \textbf{Acknowledgements.}
This work was supported by Institute of Information \& communications Technology Planning \& Evaluation (IITP) and the National Research Foundation of Korea(NRF) funded by the Korea government(MSIT) under Grants 2021-0-01176 and 2022R1A2C200494413.


% \clearpage  % TODO REVIEW/FINAL: This \clearpage needs to be removed from both review and camera-ready versions.

\bibliographystyle{splncs04}
\bibliography{egbib}
\end{document}